\documentclass{article}

\usepackage[table]{xcolor}
\usepackage[preprint]{neurips_2025}

\usepackage[utf8]{inputenc} 
\usepackage[T1]{fontenc}    
\usepackage{url}            
\usepackage{booktabs}       
\usepackage{amsfonts}       
\usepackage{nicefrac}       
\usepackage{microtype}      
\usepackage{xcolor}         

\usepackage{subcaption}

\usepackage{graphicx}
\usepackage{caption}
\usepackage{multirow}
\usepackage{titletoc}
\usepackage{enumitem} 
\usepackage{textcomp}
\usepackage{colortbl}
\usepackage{booktabs}
\usepackage{animate}

\usepackage{makecell}

\usepackage{color}

\usepackage{colortbl}
\usepackage{wrapfig}
\usepackage{csquotes}
\usepackage{amsmath}

\usepackage[ruled,vlined,linesnumbered]{algorithm2e}
\usepackage{algpseudocode}
\definecolor{Red}{RGB}{192, 0, 0}
\definecolor{Blue}{RGB}{12, 114, 186} 
\definecolor{myrefcolor}{rgb}{0, 0.367, 0.7}
\definecolor{Red}{RGB}{192, 0, 0}
\definecolor{Blue}{RGB}{12, 114, 186}
\usepackage[colorlinks=true, allcolors=myrefcolor]{hyperref}

\newcommand{\ours}{{ Follow-Your-Creation}\xspace}

\usepackage[utf8]{inputenc} 
\usepackage[T1]{fontenc}    
\usepackage{hyperref}       
\usepackage{url}            
\usepackage{booktabs}       
\usepackage{amsfonts}       
\usepackage{nicefrac}       
\usepackage{microtype}      
\usepackage{colortbl}
\definecolor{light}{gray}{0.95}
\usepackage{graphicx}   
\usepackage{booktabs}
\usepackage{array}    
\usepackage{makecell}
\usepackage[font=small, labelfont=bf, labelsep=space]{caption}
\usepackage{amsmath}
\usepackage{multirow}
\usepackage{gensymb}
\usepackage{csquotes}
\definecolor{mybrown2}{RGB}{68,63,55}

\definecolor{Red}{RGB}{192, 0, 0}
\definecolor{Blue}{RGB}{12, 114, 186} 
\definecolor{Pink}{RGB}{240,56,231}

\title{Follow-Your-Creation: Empowering 4D Creation through Video Inpainting}

\author{%
  Yue Ma$^{1}$\thanks{Equal contribution.
  $\dagger$ Corresponding author. $\ddagger$ Project leader.}, 
    Kunyu Feng$^{2\ast}$, 
  Xinhua Zhang$^{4\ast}$, 
  \textbf{Hongyu Liu$^{1\ddagger}$},
  \textbf{David Junhao Zhang$^{3}$},\\
  \textbf{Jinbo Xing$^{5}$},
  \textbf{Yinhan Zhang$^{2}$}, 
  \textbf{Ayden Yang},
  \textbf{Zeyu Wang$^{2,1\dagger}$},
  \textbf{Qifeng Chen$^{1\dagger}$}
  \vspace{.3em}
  \\$^{1}$HKUST, $^{2}$HKUST(GZ), $^{3}$NUS, $^{4}$Tsinghua Univerisity, $^{5}$CUHK \vspace{.3em} \\
   \url{https://follow-your-creation.github.io}
}

\vspace{-0.3em}

\begin{document}

\maketitle
\begin{figure}[h]

  \centering
  \includegraphics[width=1.0 \textwidth]{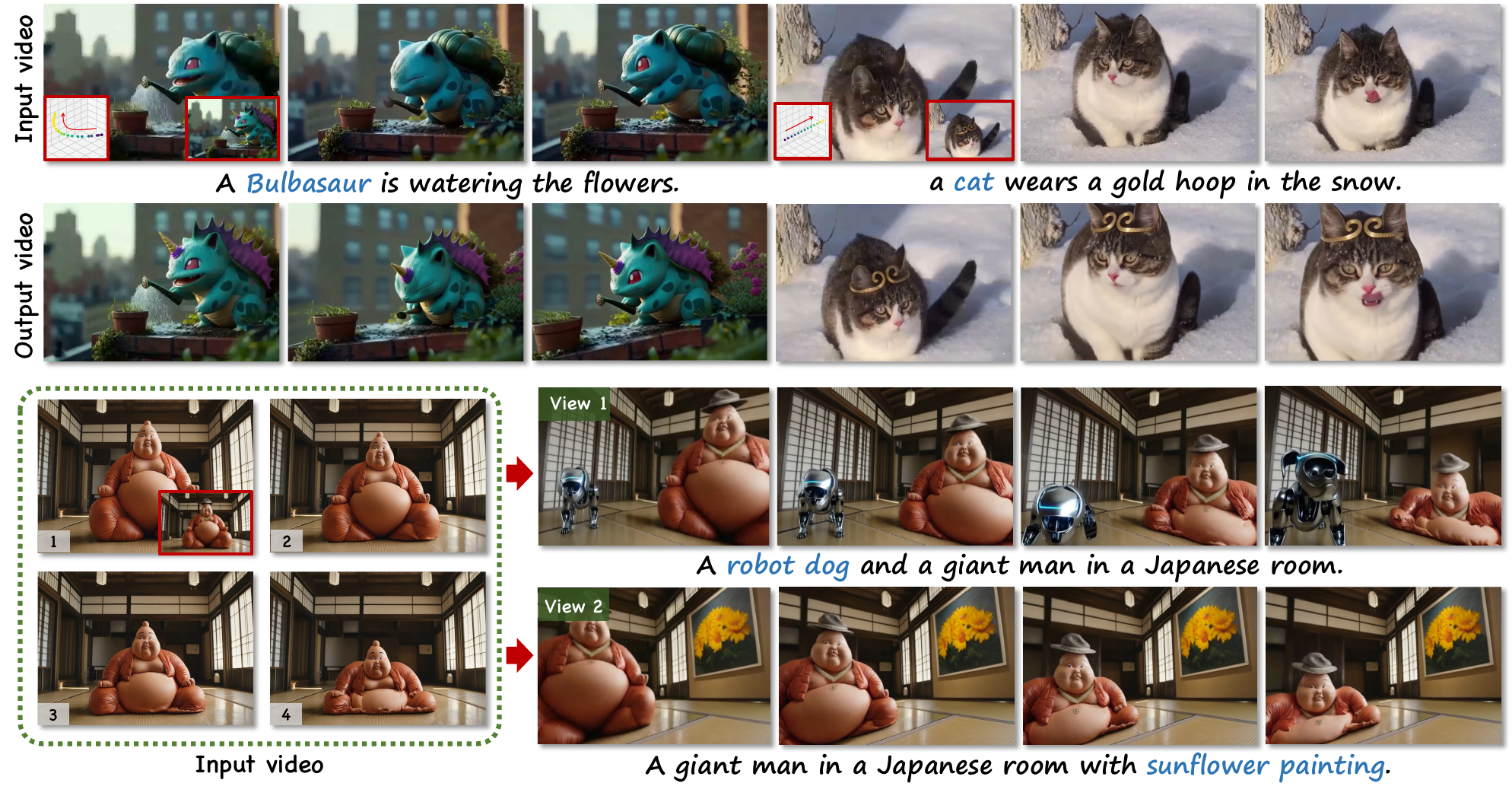} 

  \caption{\textbf{Showcases of our Follow-Your-Creation}. We reformulate 4D video creation as a video inpainting task.  Given an input video, \ours enables 4D video creation with various camera trajectories~(bottom left on the input video’s first frame) and edited first frame~(bottom right), while maintaining multi-view consistency. In addition, it supports flexible prompt-based content editing (e.g., adding a robot dog or sunflower painting).}

\label{teaser}
\end{figure}

\begin{abstract}

We introduce Follow-Your-Creation, a novel 4D video creation framework capable of both generating and editing 4D content from a single monocular video input. By leveraging a powerful video inpainting foundation model as a generative prior, we reformulate 4D video creation as a video inpainting task, enabling the model to fill in missing content caused by camera trajectory changes or user edits. To facilitate this, we generate composite masked inpainting video data to effectively fine-tune the model for 4D video generation. Given an input video and its associated camera trajectory, we first perform depth-based point cloud rendering to obtain invisibility masks that indicate the regions that should be completed. Simultaneously, editing masks are introduced to specify user-defined modifications, and these are combined with the invisibility masks to create a composite masks dataset. During training, we randomly sample different types of masks to construct diverse and challenging inpainting scenarios, enhancing the model’s generalization and robustness in various 4D editing and generation tasks. To handle temporal consistency under large camera motion, we design a self-iterative tuning strategy that gradually increases the viewing angles during training, where the model is used to generate the next-stage training data after each fine-tuning iteration. Moreover, we introduce a temporal packaging module during inference to enhance generation quality. Our method effectively leverages the prior knowledge of the base model without degrading its original performance, enabling the generation of 4D videos with consistent multi-view coherence. In addition, our approach supports prompt-based content editing, demonstrating strong flexibility and significantly outperforming state-of-the-art methods in both quality and versatility.

\end{abstract}

\section{Introduction}
Video generation foundation models~\cite{kelingAI, videoworldsimulators2024,kong2024hunyuanvideo,Wang2025WanOA} have attracted considerable attention and witnessed rapid progress in recent years. These models are capable of synthesizing high-fidelity, temporally coherent videos from coarse user inputs such as text or image prompts. As the capabilities of such models continue to evolve, there is an increasing demand for more precise and controllable video generation~\cite{he2024cameractrl, ling2024motionclone, wang2024motioninversion, 10.1007/978-3-031-72952-2_23}, where additional modalities—such as audio~\cite{chen2024echomimiclifelikeaudiodrivenportrait, hong2025audiovisualcontrolledvideodiffusion, tan2024edtalkefficientdisentanglementemotional,xu2024hallohierarchicalaudiodrivenvisual}, human pose~\cite{changmagicpose,hu2023animateanyone,ma2024follow,peng2025controlnextpowerfulefficientcontrol}, or depth~\cite{Gen-1,guo2024sparsectrl, pang2024dreamdance,xing2024make}—are leveraged to guide content creation. These advancements aim to better align the generated outputs with users' creative intent and enhance the expressiveness and relevance of the synthesized videos.

4D video generation is an emerging paradigm of controllable video synthesis that enables dynamic content creation guided by camera trajectories. It has garnered increasing attention for its ability to produce cinematic effects such as camera motion and bullet time, supporting immersive and expressive visual storytelling. Recent approaches~\cite{bahmani2024ac3d, bai2024syncammaster, fu20243dtrajmaster} typically incorporate camera trajectories into pre-trained video generation foundation models by encoding them as embeddings, analogous to text prompts. These methods often rely on multi-view datasets, synthetic renderings, or monocular videos with annotated camera poses for model fine-tuning. However, despite recent progress, several limitations remain, including a strong dependence on large-scale training data, restricted input modalities—commonly limited to images or text—and limited controllability over camera viewpoints. Notably, current methods lack support for video inputs and are unable to transform user-provided monocular videos into coherent 4D representations.

To achieve more realistic generation results and enable the conversion of monocular videos into 4D representations, recent methods~\cite{Jeong2025ReangleAVideo4V,Ren2025GEN3C3W,yu2025trajectorycrafter, zhang2024recapture} often decompose the task into two stages. The first stage employs existing depth predictors~\cite{hu2025-DepthCrafter} to estimate per-frame depth from monocular videos, generating dynamic point clouds that are rendered along desired camera trajectories. This rendering process typically results in videos with masked regions (holes) caused by occlusions or incomplete geometry. In the second stage, video inpainting is applied to fill these holes and produce the final output. While dynamic point clouds can now be reliably obtained using well-developed depth estimation techniques, appropriate video inpainting models for this task remain underdeveloped. Consequently, current approaches often rely on collecting additional data to fine-tune general-purpose image or text-to-video foundation models for inpainting. However, since these foundation models are not originally designed for video inpainting, they frequently fail to produce temporally consistent and visually realistic completions. Moreover, such fine-tuned models typically do not support text-based editing during generation, which significantly limits their flexibility and practicality for 4D video editing.

Fortunately, a powerful video inpainting foundation model—Wan2.1~\cite{Wang2025WanOA}—has recently emerged, trained on large-scale datasets. However, we observe that it cannot be directly applied to complete the masks (i.e., occluded regions) introduced by point cloud rendering, as such masks fall outside its training distribution. In this paper, we propose Follow-Your-Creation, a novel 4D video generation framework that reformulates 4D generation as a specialized video inpainting task. Our goal is to unlock the potential of powerful video inpainting models for 4D reconstruction, enabling realistic, flexible, and controllable results with minimal additional training. Specifically, we first utilize an off-the-shelf depth predictor~\cite{hu2025-DepthCrafter} to estimate per-frame depth maps, which are then transformed and aggregated into dynamic point clouds. These point clouds are rendered using a double-reprojection strategy~\cite{yu2025trajectorycrafter} to generate a sequence of masks from the target camera viewpoint, projected back to the original camera poses. These masks correspond to occluded or invisible regions caused by rendering and serve as the completion targets during 4D generation. In addition, we construct an editing mask sequence to define regions requiring content modification. The occlusion and editing masks can be used individually or combined, forming a composite mask dataset with three types of masks. During fine-tuning, we randomly sample one type of mask for each training instance, allowing the foundation model to be sufficiently trained while supporting both 4D generation and editing. Furthermore, we introduce a self-iterative tuning strategy that progressively increases viewpoint diversity by reusing results from previously trained views as training data for subsequent ones, improving stability under large camera motions.

To ensure multi-view consistency during inference, we present a temporal packing strategy that enhances coherence across frames and viewpoints by leveraging previously generated results as priors to guide content completion within masked regions. We conduct comprehensive evaluations of Follow-Your-Creation on both synchronized multi-view datasets and large-scale monocular video datasets. Quantitative results and qualitative visualizations consistently demonstrate that our method outperforms existing approaches in generating high-fidelity videos under novel camera trajectories.

\section{Related Work}

\noindent\textbf{Camera-controlled video generation.} Following the success of text-to-video generation models~\cite{bar2024lumiere,videoworldsimulators2024,chen2023videocrafter1,ho2022imagenvideo,polyak2024movie,singer2022make,wang2023lavie}, controllable video generation with additional control signals, such as pose~\cite{changmagicpose,chen2023attentive, zhang2024follow, ma2023followyourpose,xue2024follow,liu2023human}, depth~\cite{esser2023structure,xing2024make}, and sketch~\cite{ma2025magic, consisid, meng2024anidoc,xing2024tooncrafter}, has been developed to generate videos adhering to users' intentions more precisely. 
Camera motion control has been explored through motion LoRAs~\cite{blattmann2023stable,guoanimatediff,kuang2024collaborative}, enabling video generation with specific camera movement patterns. 
For finer control, a line of work has explored employing camera conditions through intrinsic and extrinsic matrix~\cite{wang2024motionctrl}, Plucker embedding~\cite{bahmani2024ac3d,bahmani2024vd3d,he2024cameractrl,li2025realcam,wang2024akira,xu2024camco,zheng2024cami2v}, background-point trajectory~\cite{feng2024i2vcontrol,wang2024motionctrl}, point-cloud re-rendering~\cite{yu2024viewcrafter}, or depth-based warping~\cite{hou2024training}.

\noindent\textbf{Diffusion-based video editing and inpainting.}
The field of video editing has broad applications. Early studies~\cite{ cao2023masactrl,hertz2022prompt2prompt,kawar2023imagic, liu2020rethinking,liu2021pd,wan2024unipaint, liu2021deflocnet,zhu2025multibooth,meng2021sdedit,mokady2023nulltextinversion, zhu2022one, liu2025avatarartist, ma2024followyourclick, ma2024followyouremoji, ma2023followyourpose, chen2024m} developed training-free or fine-tuned text-driven editing methods on images. Some
works~\cite{cong2023flatten} extend text-to-image models, where TAV\cite{wu2022tune} achieved video generation through one-shot tuning, with later works\cite{ceylan2023pix2video,ma2022visual, ma2025magic, wang2024cove, wang2024taming,zhu2024instantswap,feng2025dit4edit, zhang2025magiccolor,chai2023stablevideo, ma2025magic,xiong2025enhancing, ouyang2023codef,pumarola2021d, qi2023fatezero} improving temporal consistency. Video inpainting is a subtask of editing, which utilizes the user-specified mask sequences to edit the content in a video. Previous works can be classified into two
categories: non-generative methods and generative methods. Non-generative methods~\cite{hu2020proposal,liu2021decoupled,zhou2023propainter}  facilitate pixel propagation using architecture priors.  But they are limited to only being effective for partial object occlusions with random masks. With the development of generative models, some works~\cite{bian2025videopainter,chen2024follow, zhang2023avid,Zi2024CoCoCo, yan2025eedit} adopt the advanced text-to-video diffusion to improve their performance. They only focus on the content inpainting in video. In contrast, we reformulate 4D video creation as a video inpainting task and unlock the potential of video inpainting models for 4D reconstruction, enabling high-fidelity results with minimal additional training.

\noindent\textbf{Novel view synthesis of dynamic scenes.} With the rapid development of diffusion models in the image and video generation domain, pre-trained video diffusion models have demonstrated strong capabilities in novel view generation. AvatarArtist~\cite{liu2025avatarartist} leverages diffusion models to predict a 4D representation of avatars, but it is not well-suited for general scenes.  DimensionX~\cite{sun2024dimensionx} integrates a dedicated motion LoRA for dynamic new view synthesis, but its camera motion is limited to a few simple trajectories.
RecamMaster~\cite{Bai2025ReCamMasterCG} and TrajectoryCrafter~\cite{yu2025trajectorycrafter} employ the large-scale video dataset to tune the text-to-video diffusion model, which incurs significant computational costs.  
Some works~\cite{Jeong2025ReangleAVideo4V,zhang2024recapture} use masked loss and regenerate videos using point cloud rendering and mask fine-tuning along custom camera paths. 
Despite having LoRA adaptation capabilities, they struggle with generating 4D videos with large camera motion.
In contrast, our approach can generate editable, high-quality 4D videos with multi-view consistency over a larger angle range.

\section{Method}
\begin{figure*}[t]
  \centering
  \includegraphics[width=\linewidth]{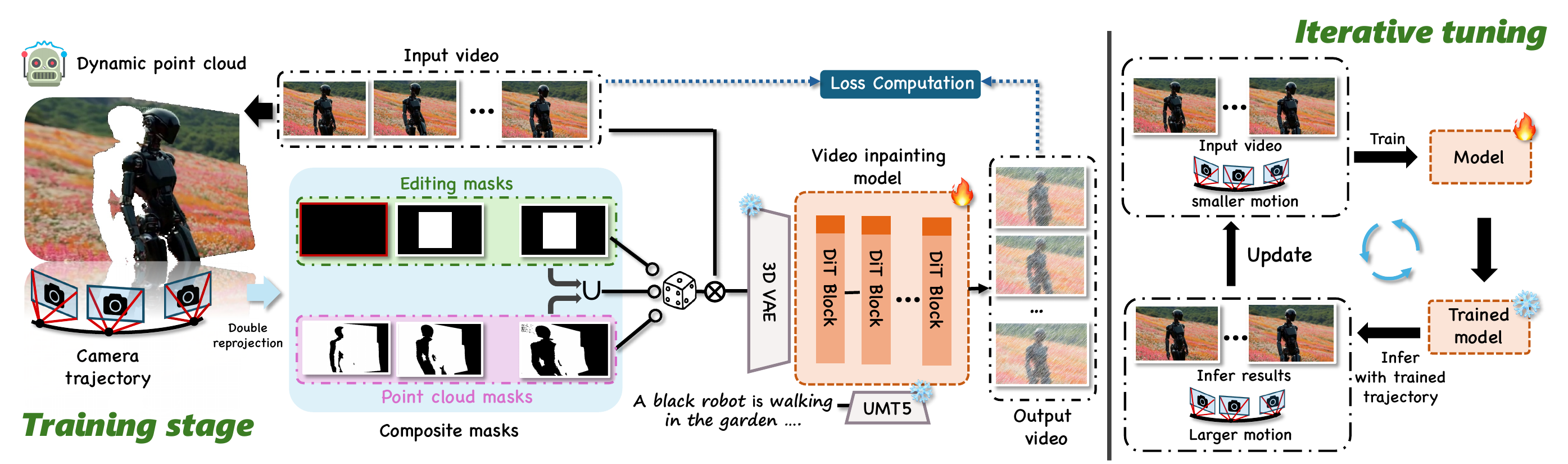}
  \caption{\textbf{Overview of our method.} We reformulate the 4D video creation as video inpainting task. \textbf{Left:} given a video, we first generate the composite masks from the dynamic point cloud and feed them into video inpainting model to unlock its 4D video creation capability. \textbf{Right:} To unlock the capability of generating 4D video with larger motion, we first generate videos with small motion, then feed them
  into the model to improve the temporal consistency progressively.
  }
  \label{fig:framework}
\end{figure*}

The pipeline of our method is shown in Fig.~\ref{fig:framework}. Given a video, we target the creation of dynamic 4D video content, including camera retargeting (\textit{e.g.}, zoom, tilt, pan) and video content editing (\textit{e.g.}, subject addition and modification). Different from previous works~\cite{Bai2025ReCamMasterCG, yu2025trajectorycrafter} tuned on
large-scale video datasets, we reformulate 4D generation as a specialized video inpainting task and fine-tune the model~\cite{Wang2025WanOA} to unlock its potential. In the following, we first discuss the details about the dynamic point cloud  (in Sec.~\ref{sec:dynamicpointcloud}) and composite mask~(in Sec.~\ref{sec:Composite mask}). Then we introduce our iterative tuning in Sec.~\ref{sec:Iterative tuning}. Finally, we present the temporal-packing inference for multi-view video consistency in Sec.~\ref{sec:Temporal-packing inference}.  



\subsection{Dynamic point cloud}
\label{sec:dynamicpointcloud}

Given an input video $\mathbf{V} = [\mathbf{I}_{0},\dots,\mathbf{I}_{N-1}] \in \mathbb{R}^{N \times 3 \times H \times W}$, where $N$ denotes the number of frames, our goal is to synthesize a new video sequence that follows a user-specified camera trajectory. The dynamic point cloud serves as a crucial intermediary representation that bridges the original frames and the novel camera views. Specifically, we use the off-the-shelf video depth estimator~\cite{hu2025-DepthCrafter} to estimate per-frame depth map $\mathbf{D} = \{\mathbf{D}_{i}\}_{i=0}^{N-1}$. 
By combining the video frames and their depth maps, the point cloud sequence $\mathcal{P}=\{\mathbf{P}_{i}\}$ can be computed as follows:
\begin{equation} 
\mathcal{P}_i = \phi([\mathbf{I}_i, \mathbf{D}_i], \mathbf{K}),
\end{equation}
where $\phi$ is a function that maps $[\mathbf{I}_i, \mathbf{D}_i]$ to a 3D point cloud in the camera's coordinate system using $\mathbf{K}$ that represents the intrinsics of the camera described in~\cite{chung2023luciddreamer, Jeong2025ReangleAVideo4V}.  Additionally, the camera motion is provided as a sequence of extrinsic matrices $\mathcal{T} = \left \{\mathbf{T}_{1}, \ldots, \mathbf{T}_{N-1} \right \}$. 
With these extrinsic matrices, we can project the point cloud from each frame back to the camera plane using the perspective function 
$\psi$ to render an image:
\begin{equation}
\mathbf{I}^{a}_{i} = \psi (\mathcal{P}_i, \mathbf{K}, \mathbf{T}_i).
\end{equation}

However, the rendered results are often incomplete, as a single monocular depth map is insufficient to reconstruct the entire scene, leading to occluded or missing regions. These missing areas can be identified during the rendering process, where a binary visibility mask $\mathbf{M} \in \mathbb{R}^{N \times 1 \times H \times W}$ is generated. Pixels with valid projections are marked as 1, while regions falling outside the original view due to camera motion are marked as 0. Our method aims to leverage a video inpainting foundation model to fill in such masked regions, thereby enabling the generation of a complete 4D video.

\subsection{Composite mask}
Apart from the aforementioned visibility masks, our method also supports 4D editing, which requires an additional mask to indicate the user-specified editing regions. In the following sections, we describe how we construct various types of masks and integrate them into a composite mask dataset that serves as the supervision for training our video inpainting model.
\label{sec:Composite mask}

\noindent\textbf{Point cloud mask.} 
It is hard to directly use the original visibility masks in in Sec.~\ref{sec:dynamicpointcloud}  for supervision, since the rendered frame $\mathbf{I}^{a}_{i}$ does not contain ground-truth content for the occluded regions. To overcome this, we adopt a double reprojection strategy~\cite{yu2025trajectorycrafter} that reprojects the visibility masks back to the viewpoint of the input video. This allows us to obtain supervision masks that are spatially aligned with the original frames, and we can set the input video as the ground truth for training of the inpainting model. Specifically,  we back-project the rendered views $\mathbf{V}^{\prime}$ into a new point cloud $\mathcal{P}^{\prime}$ using $\mathbf{D}^{\prime}$ and re-render the view $\mathbf{V}^{\prime\prime}$ by applying the inverse transformation $\mathcal{T}^{-1}$. This results in a paired set consisting of the corrupted video $\mathbf{V}^{\prime\prime}$ with the Mask $\mathbf{M}^{\prime\prime}$ to indicate the artifacts region and the corresponding clean video $\mathbf{V}^s$, both following the same original camera trajectory.



\noindent\textbf{Editing mask.} For 4D video generation, content editing plays a crucial role in practical applications. To leverage powerful inpainting priors and enable more flexible video editing, we adopt a content editing strategy inspired by prior work~\cite{mou2025revideo}. During training, we randomly select a region and generate the corresponding mask sequence. The mask for the first frame is set to '0', indicating that the first frame serves as the guidance for video synthesis. During inference, content editing is performed by modifying the first frame, and the changes are subsequently propagated to the following frames.

\noindent\textbf{Union mask.} To enable both 4D completion and 4D editing tasks simultaneously, we combine the aforementioned two types of masks using a union operation to form a new composite mask.

After obtaining various video-mask pair data, we randomly select three types of masks: point cloud masks, editing masks, and their union in the training stage. The conventional diffusion inpainting pipeline is adopt, which takes the corrupted video $\mathbf{V}^{\prime\prime}$ and the corresponding occluded mask $\mathbf{M}^{\prime\prime}$ as input conditions and predicts the completed video $\mathbf{V}^{s}$ using a standard flow matching loss \cite{Wang2025WanOA}. 

\subsection{Self-iterative tuning}  
\label{sec:Iterative tuning}

 By leveraging a powerful video inpainting foundation model as a generative prior, we regard the 4D video creation as a video inpainting task. However, a vanilla video inpainting diffusion model has a challenge in handling the hole video with a larger angle (\textit{e.g.}, $>$ 40 degrees).  
 The reason behind this lies in two key factors. 
 Firstly, in our setting, the video inpainting methods typically perform fine-tuning on a single video with various masks, which restricts their generalization ability. Secondly, current video inpainting models lack robust 3D perception capabilities. Consequently, these methods struggle to generate large-angle scene reconstructions in videos while maintaining temporal consistency across frames. To generate the video with larger-angle content, we propose the self-iterative tuning, which enhances the 3D generation ability of the video inpainting model progressively.

In detail, given a reference video $\mathbf{V} = [\mathbf{I}_{0},\dots,\mathbf{I}_{N-1}] \in \mathbb{R}^{N \times 3 \times H \times W}$ as defined in Sec.~\ref{sec:dynamicpointcloud}, our pipeline initiates by generating multiple video-mask pairs $\{(\mathbf{V}^{(k)},\mathbf{M}^{(k)})\}_{k=1}^{K}$ using small viewpoints (\textit{e.g.}, < 30 degrees). 
These pairs are used to perform one-shot tuning of the video inpainting model through Low-Rank Adaptation (LoRA), and optimized parameters 
\begin{equation}
\mathbf{W}_{LoRA}^* = \arg\min_{\mathbf{W}} \mathcal{L}\left(\mathbf{V}^k,\mathbf{M}^k,  \Delta\mathbf{W}\right),
\end{equation}
where $\Delta\mathbf{W}$ denotes the low-rank parameter updates. After tuning, we load the LoRA weight $\mathbf{W}_{LoRA}^*$ and infer the video $\widetilde{\boldsymbol{{V}}}$ with trained small angles. Subsequent iterations employ an angle-progressive scheme: at each loop $j$, we generate a new masked video with larger angular ranges, The model then performs a self-iterative pipeline through the recurrence relation:

\begin{gather}
\widetilde{\mathbf{I}}^{j}_i = \psi\left( \mathcal{P}_{i}^{j}, \mathbf{K}, \mathbf{T}^{j}_{i} \right), \quad i = 0, 1, \dots, N-1 \label{eq1} \\
\widetilde{\boldsymbol{V}}^j = \left\{ \mathbf{I}_0^j, \dots, \mathbf{I}_{N-1}^j \right\} \label{eq2} \\
\mathbf{W}_{LoRA}^{(j)} = \mathbf{W}_{LoRA}^{(j-1)} + \eta \nabla_{\mathbf{W}} \mathcal{L}_{cycle} \left( \widetilde{\boldsymbol{V}}^j, \mathbf{M}^j ,\Delta\mathbf{W}\right) \label{eq3}
\end{gather}
where $\psi(\cdot )$ denotes our geometric warping function that extrapolates viewpoints. The $\eta $ is the learning rate and $\mathcal{L}_{{cycle}}$ means spatial-temporal consistency MSE losses~\cite{hastie2009elements}. 

\subsection{Temporal-packing inference}
\label{sec:Temporal-packing inference}%
\begin{wrapfigure}{r}{0.5\textwidth}
  \centering
  \vspace{-5mm}
  \includegraphics[width=0.5\textwidth]{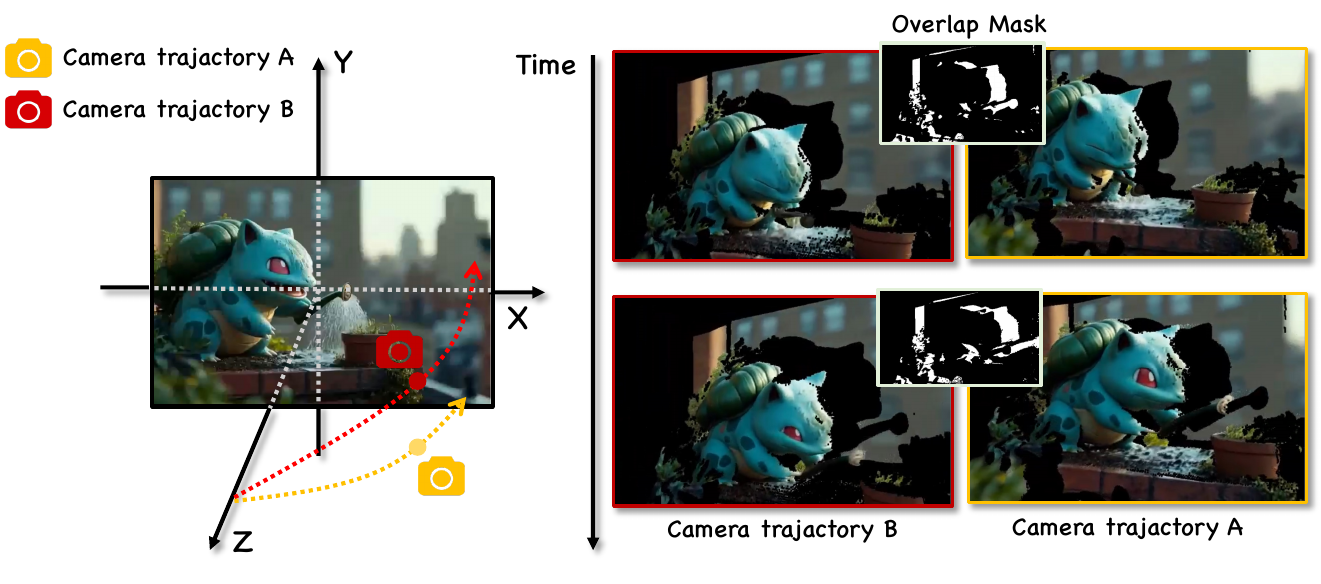} 
    \captionof{figure}{\textbf{Motivation of temporal-packing inference}. During the generation, there are existing overlaps~(overlap mask) between various camera poses, which enables improving the consistency in multiple views. } 
    \label{fig:motivation}
\end{wrapfigure}

 After finishing the self-iterative tuning,  we aim to achieve 4D video creation using various camera trajectories and the edited first frames. However, there is still a challenge in 4D video creation: multi-view video consistency. In detail, our goal is to preserve the subject and scene consistency in generated multi-view videos.  Previous works, such as Recaputre~\cite{zhang2024recapture}, RecamMaster~\cite{Bai2025ReCamMasterCG}, and TrajactoryCrafter~\cite{yu2025trajectorycrafter}, only focus on the consistency between the input and generated videos rather than multi-view generated videos.
 Reangle-a-video~\cite{Jeong2025ReangleAVideo4V} utilizes the image inpainting tools to improve the consistency. 
 But it requires manual selection. In our work, during inference stage, we propose the temporal-packing strategy to maintain the multi-view video consistency. Specifically, as shown in Fig.~\ref{fig:motivation}, we first observe that the rendered hole videos from two different camera trajectories have the overlap areas (overlap mask in Fig.~\ref{fig:motivation}). Inpainting the same area in two forwards will lead to regional inconsistency.  To improve the coherency of multi-view video, after obtaining the generated video $\widetilde{\boldsymbol{V}}^a$ from camera trajectory  $\mathcal{T}^{a}$ using our \ours, then we calculate the area of inpainting in each frame and select the $K$ frames in $\widetilde{\boldsymbol{V}}^a$, 
\begin{equation}
    \mathbf{F} = \text{top-$k$-argmax}(\boldsymbol{S}[\widetilde{\boldsymbol{V}}^a,\mathbf{M}^{a}]),
\end{equation}  
where $\boldsymbol{S}(\cdot)$ notes area calculation function. In the next inference for camera trajectory $\mathcal{T}^{b}$,  we concatenate the selected frames' tokens with the hole video token along the temporal dimension: 
\begin{equation}
\begin{aligned}
    x_{input} &= [\texttt{patchify}(\mathcal{E}(\mathbf{F})), \texttt{patchify}(\mathcal{E}(\mathbf{\mathbf{V}}^{b}))]_{\text{temporal}}, \\
\end{aligned}
\end{equation}
where $x_{input} \in \mathbb{R}^{B \times 2N \times S \times C}$ is the input of video inpainting model, and $S = H \times W$, $C$ is the channel dimension for the latent diffusion model. $\mathcal{E}(\cdot)$ notes pretrained 3D-VAE~\cite{kong2024hunyuanvideo}. Note that 
we do not design any additional attention layers for feature fusion. In a pretrained video inpainting model~\cite{Wang2025WanOA}, self-attention is applied globally across all tokens within the spatio-temporal attention layers. 

\section{Experiments}

\begin{figure*}[t]
  \centering
  \includegraphics[width=\linewidth]{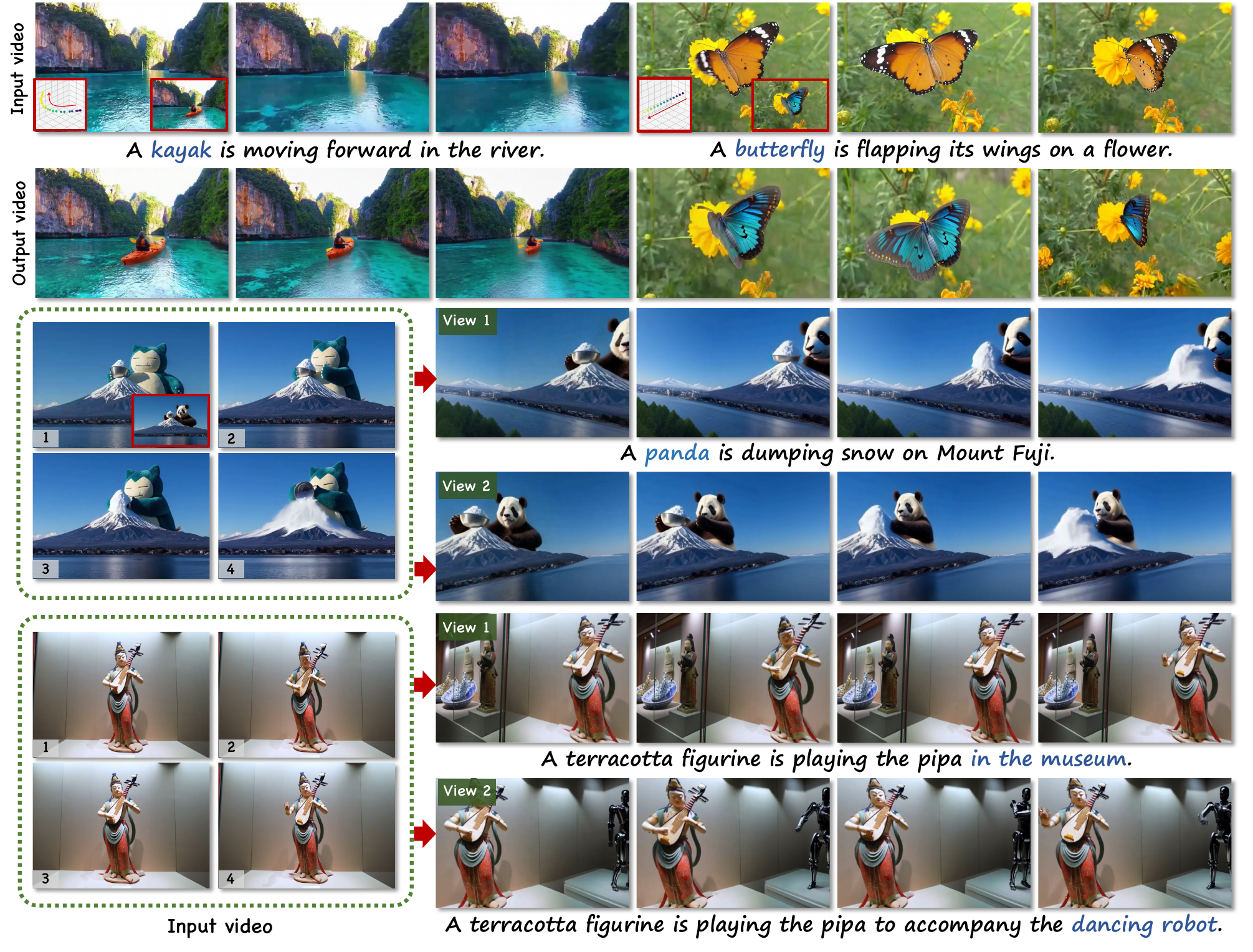}
  \caption{\textbf{Gallery of our proposed method.} Our Follow-Your-Creation enables achieving flexible and high-quality 4D video creation using the given camera trajectory and the edited first frame~(2nd row). Additionally, it also supports the 4D video creation using various prompts in frozen camera~(\enquote{exhibition} in 4th row and \enquote{robot} in 6th row).   }
  \label{fig:gallery4}
  \vspace{-0.6cm}
\end{figure*}

\subsection{Implementation details}
In our experiment,  the open-sourced video generation model WAN-2.1~\cite{Wang2025WanOA} is employed as the base text-to-video generative model. We use the LoRA~\cite{Hu2021LoRALA} to finetune the model and the ranks are 128. During one-shot training, each video is input into the model as $512 \times 512$, and the video length is set to 81. The training stage is conducted for 2000 steps with learning rate $1 \times 10^{-5}$ and weight decay $0.1$. For producing dynamic point cloud, depth sequences are evaluated from input video using the open-sourced depth estimator DepthCrafter~\cite{Hu2024DepthCrafterGC}, with empirically configured camera intrinsics.  We optimize our model using PyTorch on a single NVIDIA A800 GPU for about 2 hours. During inference, we employ the DPM solver~\cite{Lu2022DPMSolverAF} with 30 sampling steps and a text-guidance scale of 6.5. The LoRA weights are fixed at 0.7. Additional implementation details and evaluation metrics are provided in the supplementary materials.

\begin{figure*}[t]
  \centering
  \includegraphics[width=\linewidth]{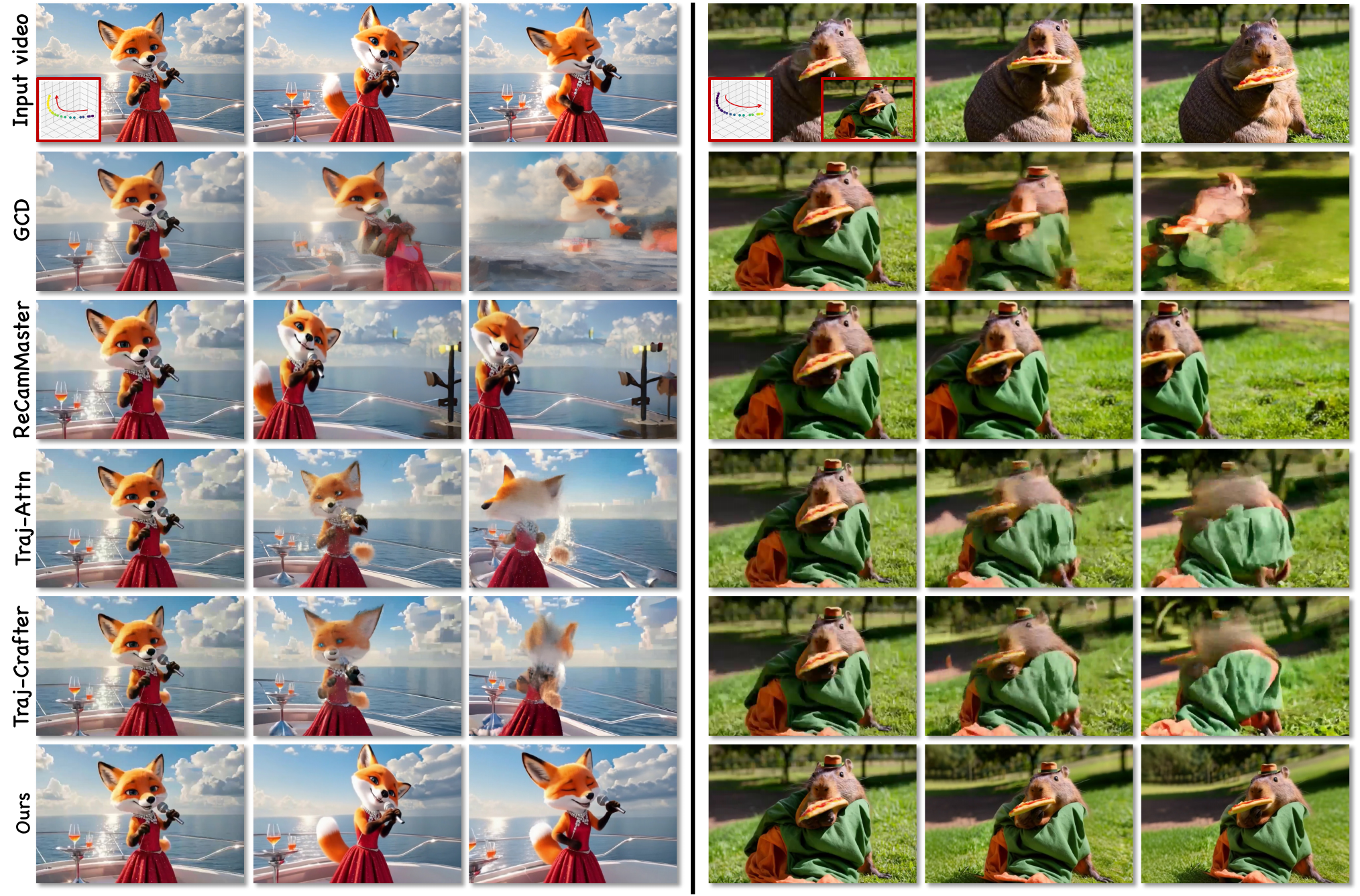}
  \caption{\textbf{Qualitative comparison results with the state-of-the-art methods.}  The results show that our Follow-Your-Creation exhibits 4D video creation with better consistency and camera movements.}
  \label{fig：compare}
\end{figure*}

\subsection{Comparison with baselines}

\noindent\textbf{Qualitative comparison.} 
We first evaluate the camera motion editing ability of our method with four baselines: Trajectory-Attention~\cite{Xiao2024TrajectoryAF}, ReCamMaster~\cite{Bai2025ReCamMasterCG}, TrajectoryCrafter~\cite{yu2025trajectorycrafter}, GCD~\cite{Hoorick2024GenerativeCD}. The first method is a diffusion-based novel view synthesis method, and ReCamMaster and TrajactoryCrafter are state-of-the-art generative camera-retargeting methods tuned on a large-scale video dataset.  
GCD is a 4D novel view synthesis technique, which integrates implicit camera pose embeddings into a video diffusion model.
The visual comparison is shown in Fig.~\ref{fig：compare}(left). We can see that videos generated by GCD exhibit over-smoothed details and view misalignment issues. 
On the other hand, while TrajactoryCrafter and ReCamMaster demonstrate better pose accuracy, they struggle to produce high-fidelity frames. In contrast, our method is capable of generating novel trajectory videos with high fidelity, remarkable 4D consistency, and precise pose control.

Additionally, to test the ability of 4D video creation, we conduct a comparison with a baseline approach. 
In detail, we first edit the video using an advanced video editing tool~\cite{jiang2025vace}, then feed it into the camera-retargeting baselines for visual comparison.
In Fig.~\ref{fig：compare}(right), we present the visual quality of the results produced by our method and the baselines. It's observed that the 4D video content generated by baseline methods has significant artifacts. On the other hand, our method is capable of achieving better editing effects for 4D videos, along with smooth and accurate camera movements.

\begin{table*}[t]
\setlength{\tabcolsep}{3pt} 
\centering
\small 

\caption{
  \textbf{VBench results between ours and baselines.}
  We collect a comprehensive video benchmark with 40 real-world videos and 40 high-quality generated videos to evaluate the performance. 
  \textcolor{Red}{\textbf{Red}} stands for the best result, \textcolor{Blue}{\textbf{Blue}} stands for the second best result.
}
\label{tab:vbench}
\resizebox{0.98\textwidth}{!}{
\begin{tabular}{@{} l *{5}{c} @{}} 
\toprule
& \multicolumn{5}{c}{VBench $\uparrow$} \\
\cmidrule(lr){2-6}
Method & Subject Consis.$\uparrow$ & Background Consis.$\uparrow$ & Temporal Flicker.$\uparrow$ & Motion Smooth.$\uparrow$ & Overall Consis.$\uparrow$ \\
\midrule

GCD~\cite{Hoorick2024GenerativeCD}             & 0.7245          & 0.7438             & 0.6984           & 0.7041        & 0.1932          \\
Trajectory-Attention~\cite{Xiao2024TrajectoryAF} & 0.7419          & 0.7821             & 0.7346           & 0.7528        & 0.2087          \\
ReCamMaster~\cite{Bai2025ReCamMasterCG}      & 0.8217          & 0.8437             & \textcolor{Blue}{\textbf{0.8219}}           & 0.8523        & 0.2376          \\
TrajectoryCrafter~\cite{Mark2025TrajectoryCrafterRC} & \textcolor{Blue}{\textbf{0.8632}}          & \textcolor{Blue}{\textbf{0.8674}}             & 0.7925           & \textcolor{Blue}{\textbf{0.8815}}        & \textcolor{Blue}{\textbf{0.2463}}          \\
\midrule
Ours                 & \textcolor{Red}{\textbf{0.9026}} & \textcolor{Red}{\textbf{0.8931}}    & \textcolor{Red}{\textbf{0.8818}}  & \textcolor{Red}{\textbf{0.9242}} & \textcolor{Red}{\textbf{0.2915}}  \\
\bottomrule

\end{tabular}%
}
\end{table*}

\begin{table*}[t!]
	\begin{center}
            \vspace{-0.30cm}
		\caption{\textbf{Quantitative comparison with state-of-the-art methods.} We perform the assessment on visual quality, camera accuracy, and view synchronization. \textcolor{Red}{\textbf{Red}} stands for the best result, \textcolor{Blue}{\textbf{Blue}} stands for the second best result.}
		\setlength\tabcolsep{3.2pt}
            \resizebox{\textwidth}{!}{
    		\begin{tabular}{lcccc|cc|ccc}
    			\toprule
    			\multirow{2}{*}{Method} & \multicolumn{4}{c|}{Visual Quality} & \multicolumn{2}{c|}{Camera Accuracy} & \multicolumn{3}{c}{View Synchronization}\\ 
                    \cmidrule(r){2-10}
                    & \makecell[c]{FID $\downarrow$} & \makecell[c]{FVD $\downarrow$} & \makecell[c]{CLIP-T $\uparrow$} & \makecell[c]{CLIP-F $\uparrow$} & RotErr $\downarrow$& TransErr $\downarrow$ & {Mat. Pix.(K) $\uparrow$} & {FVD-V $\downarrow$} & {CLIP-V $\uparrow$}\\
                    \midrule 
                    GCD~\cite{Hoorick2024GenerativeCD} & 73.92 & 368.44 & 32.81 & 93.66 & 2.25 & 5.78 & 638.76 & 364.28 & 85.94 \\
                    Trajectory-Attention~\cite{Xiao2024TrajectoryAF} & 70.33 & 275.84 & 33.08 & 94.51 & 2.15 & 5.65 & 620.17 & 239.15 & 88.53 \\
                    ReCamMaster~\cite{Bai2025ReCamMasterCG} & 64.82 & 162.91 & 34.68 & \textcolor{Blue}{\textbf{96.24}} & 1.48 & 5.58 & 628.45 & \textcolor{Blue}{\textbf{153.29}} & 88.27 \\
                    TrajectoryCrafter~\cite{yu2025trajectorycrafter} & \textcolor{Blue}{\textbf{61.57}} & \textcolor{Blue}{\textbf{154.23}} & \textcolor{Blue}{\textbf{35.27}} & 96.15 & \textcolor{Blue}{\textbf{1.43}} & \textcolor{Blue}{\textbf{5.52}} & \textcolor{Blue}{\textbf{635.25}} & 148.71 & \textcolor{Blue}{\textbf{87.42}} \\
                    \midrule
                    Ours & \textcolor{Red}{\textbf{\textbf{58.26}}} & \textcolor{Red}{\textbf{\textbf{145.71}}} & \textcolor{Red}{\textbf{\textbf{35.63}}} & \textcolor{Red}{\textbf{\textbf{96.62}}} & \textcolor{Red}{\textbf{\textbf{1.37}}} & \textcolor{Red}{\textbf{\textbf{4.47}}} & \textcolor{Red}{\textbf{\textbf{705.34}}} & \textcolor{Red}{\textbf{\textbf{119.52}}} & \textcolor{Red}{\textbf{\textbf{89.87}}} \\
    			\bottomrule
    		\end{tabular}
            \label{tab:othermetrix}
            }
	\end{center}
\end{table*}

\noindent\textbf{Quantitative comparison.} 
We perform three comprehensive quantitative assessments of the results obtained by our proposed method and the baseline. The user study is provided in the appendix.

(1) \textbf{Low-level metrics} in the Kubric-4D dataset~\cite{zhang2024recapture}.
The video has a resolution of $576 \times 384$ and spans across $60$ frames at $24$ FPS.  We select the PSNR, LPIPS, and SSIM as low-level metrics to evaluate similarity between generated and ground truth novel views. The results are reported in Tab.~\ref{tab:lowlevel}.  
The results clearly indicate that our method outperforms the baseline across all metrics.

(2) \textbf{VBench metrics}: 
We collect 40 real-world videos and 40 high-quality generated videos by advanced text-to-video generative models~\cite{kong2024hunyuanvideo, Wang2025WanOA}. For each video, we generate 5 different novel trajectory videos.  Five metrics in VBench~\cite{Huang2023VBenchCB} are employed for a more accurate evaluation (in Tab.~\ref{tab:vbench})

\begin{wrapfigure}{r}{0.5\textwidth}  
  \centering

  \renewcommand{\arraystretch}{1.2}
  \definecolor{Red}{RGB}{192,0,0}
  \definecolor{Blue}{RGB}{12,114,186}
  \begin{minipage}{0.5\textwidth}
\captionof{table}{\textbf{Comparison results on Kubric-4D.} \textcolor{Red}{\textbf{Red}} and \textcolor{Blue}{\textbf{Blue}} denote the best and second best results.}
\label{tab:kubric_sota}
\raggedright 
\resizebox{\linewidth}{!}{
\renewcommand{\arraystretch}{1.1} 
\begin{tabular}{@{}l|ccc@{}}
\toprule
Method & \thead{PSNR $\uparrow$} & \thead{SSIM $\uparrow$} & \thead{LPIPS $\downarrow$} \\
\midrule
GCD~\cite{Hoorick2024GenerativeCD} & 14.21 & 0.398 & 0.612 \\
Trajectory-Attention~\cite{Xiao2024TrajectoryAF} & 14.65 & 0.426 & 0.587 \\
ReCamMaster~\cite{Bai2025ReCamMasterCG} & 15.03 & 0.453 & 0.561 \\
TrajectoryCrafter~\cite{Mark2025TrajectoryCrafterRC} & \textcolor{Blue}{\textbf{15.82}} & \textcolor{Blue}{\textbf{0.487}} & \textcolor{Blue}{\textbf{0.532}}  \\
\midrule
Ours & \textcolor{Red}{\textbf{22.15}} & \textcolor{Red}{\textbf{0.523}} & \textcolor{Red}{\textbf{0.381}} \\
\bottomrule
\end{tabular}%
}
\label{tab:lowlevel}
\end{minipage}

  \vspace{1ex}

\vspace{-6mm}

\end{wrapfigure}

(3) \textbf{Other metrics}: following the previous work~\cite{Bai2025ReCamMasterCG}, we calculate the other metrics in Tab.~\ref{tab:othermetrix}. 
 We assess camera trajectory accuracy by calculating rotation and translation errors, following methods from earlier research in camera-guided generative approaches ~\cite{he2024cameractrl, wang2024motionctrl}. For view synchronization, we computed clip similarity scores and FVD between video frames from different viewpoints in the same scene, which we refer to as CLIP-V and FVD-V. Notably, Follow-Your-Creation outperforms baselines across multiple metric dimensions, demonstrating its superior generative consistency and visual fidelity.

\begin{table*}[t]
	\begin{center}
            \vspace{-3mm}
		\caption{\textbf{Quantitative ablation results}. \textcolor{Red}{\textbf{Red}} and \textcolor{Blue}{\textbf{Blue}} denote the best and second best results.}
            \vspace{-0.30cm}
		\label{tab_quality_eval}
		\setlength\tabcolsep{3.2 pt}
            \resizebox{\textwidth}{!}{
    		\begin{tabular}{lcccc|cc|ccc}
    			\toprule
    			\multirow{2}{*}{Method} & \multicolumn{4}{c|}{Visual Quality} & \multicolumn{2}{c|}{Camera Accuracy} & \multicolumn{3}{c}{View Synchronization}\\ 
                    \cmidrule(r){2-10}
                    & \makecell[c]{FID $\downarrow$} & \makecell[c]{FVD $\downarrow$} & \makecell[c]{CLIP-T $\uparrow$} & \makecell[c]{CLIP-F $\uparrow$} & RotErr $\downarrow$& TransErr $\downarrow$ & {Mat. Pix.(K) $\uparrow$} & {FVD-V $\downarrow$} & {CLIP-V $\uparrow$}\\
                    \midrule 
                    W/o composite mask tuning & 78.27 & \textcolor{Blue}{\textbf{153.28}} & 30.81 & 94.21 & 1.56 & 5.26 & 518.21 & 155.47 & \textcolor{Blue}{\textbf{85.25}} \\
                    W/o iterative tuning & 86.29 & 197.24 & \textcolor{Red}{\textbf{36.58}} & 92.74 & \textcolor{Blue}{\textbf{1.49}} & 4.93 & \textcolor{Blue}{\textbf{589.29}} & 204.81 & 81.26 \\
                    W/o temporal pack strategy & \textcolor{Blue}{\textbf{62.46}} & 168.91 & 34.92 & 93.44 & 1.51 & \textcolor{Blue}{\textbf{4.52}} & 524.63 & \textcolor{Blue}{\textbf{137.64}} & 84.71 \\
                    \midrule
                    Ours & \textcolor{Red}{\textbf{58.26}} & \textcolor{Red}{\textbf{145.71}} & \textcolor{Blue}{\textbf{35.63}} & \textcolor{Red}{\textbf{96.62}} & \textcolor{Red}{\textbf{1.37}} & \textcolor{Red}{\textbf{4.47}} & \textcolor{Red}{\textbf{705.34}} & \textcolor{Red}{\textbf{119.52}} & \textcolor{Red}{\textbf{89.87}} \\
    			\bottomrule
    		\end{tabular}
            \label{tab:ab_about_model}
            }
	\end{center}
        \vspace{-0.5cm}
        
\end{table*}

\begin{figure*}[t]
  \centering
  \includegraphics[width=\linewidth]{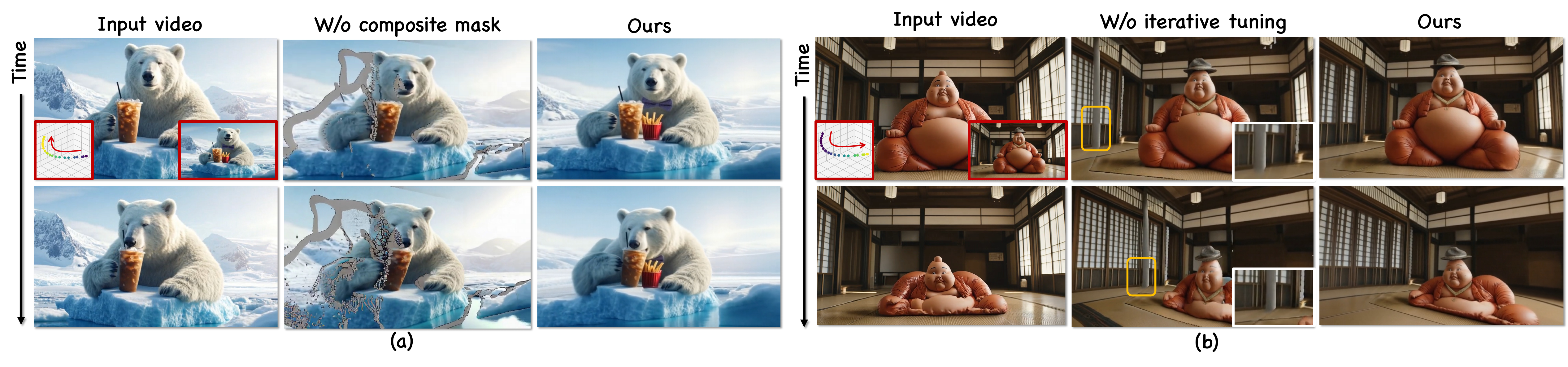}
  \vspace{-0.6cm}
  \caption{\textbf{Ablation study about composite mask (a) and self-iterative tuning (b).} Fig.~\ref{fig:abl1}~(a) demonstrates that our proposed composite mask not only keeps a smooth camera trajectory during generation process but also enables the performance of editing task. Fig.~\ref{fig:abl1}~(b) shows that the self-iterative tuning facilitates to maintain a better temporal coherence in a larger camera motion angle. }
  \vspace{-0.6cm}
  \label{fig:abl1}
\end{figure*}



\subsection{Ablation study}
\label{sec:ablation}

\noindent\textbf{Effectiveness of composite mask.} 
To investigate the contribution of composite mask during training, we conduct a series of ablation studies on the it. The experimental settings are the same during the ablation. As shown in Fig.~\ref{fig:abl1}(a), when we remove the composite mask during the training stage, the results have artifacts and fail to follow the edited first frame (\enquote{French fries} on the ice). In contrast, our method enables creating a reasonable video with the subject while following the given camera trajectories. Additionally, we also perform a quantitative ablation study~(in Tab.~\ref{tab:ab_about_model}). Without composite mask strategy,  our method fails to achieve both the 4D video creation and editing. 


\begin{wrapfigure}{r}{0.5\textwidth}
  \centering
  \vspace{-6mm}
 \begin{minipage}{0.5\textwidth}
    \centering
    \includegraphics[width=\linewidth]{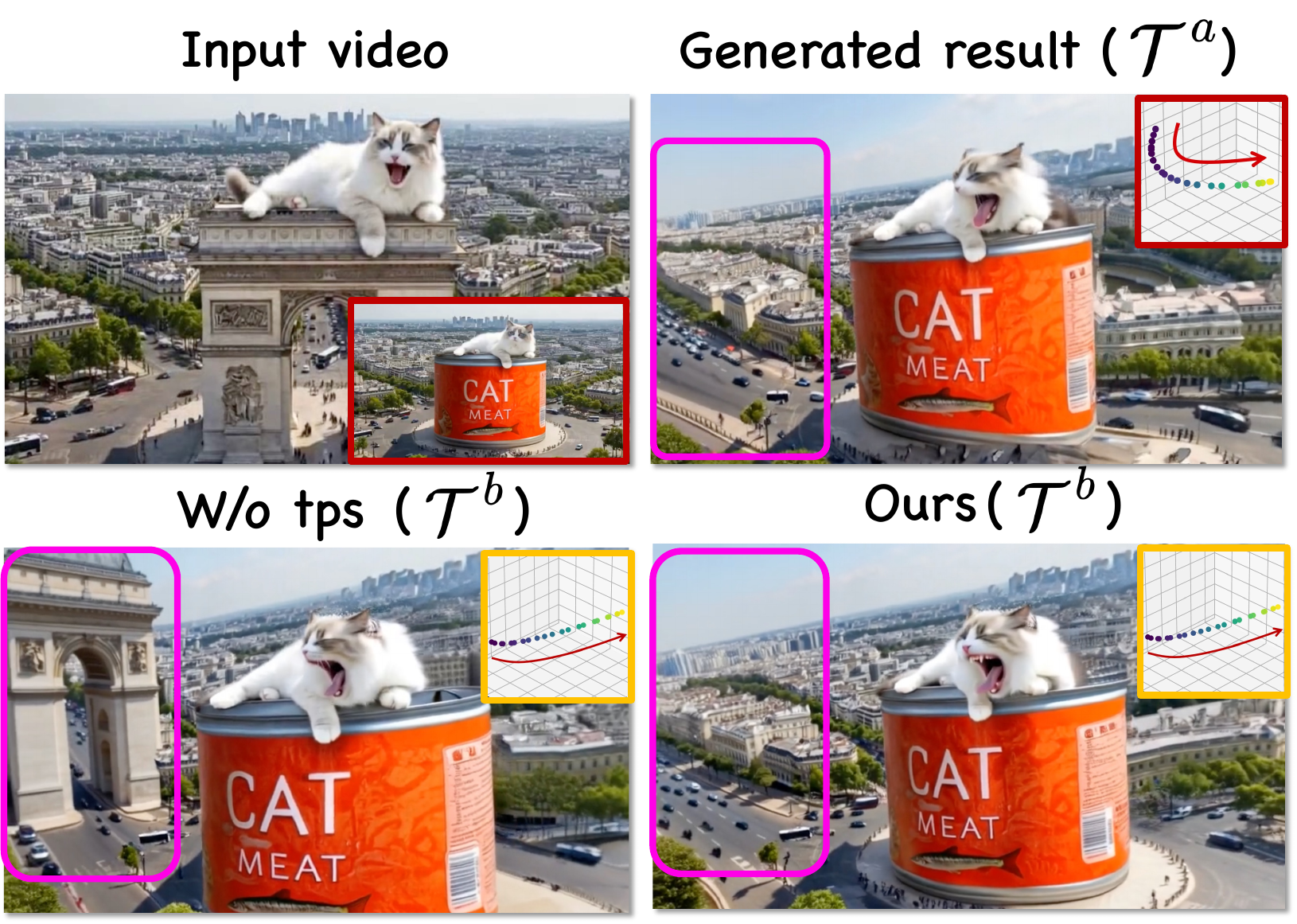} 
    \captionof{figure}{\textbf{Ablation study of temporal-packing inference.} $\mathcal{T}^{a}$ represents a rightward camera motion, while $\mathcal{T}^{b}$ turns both upward and rightward, exhibiting partial spatial overlap with $\mathcal{T}^{a}$. 
    Under the temporal-packing inference strategy, our method keeps a better multi-view consistency between two trajectories.
    }  
    \label{fig:ablation2}
\end{minipage}
\end{wrapfigure}

\noindent\textbf{Effectiveness of self-iterative tuning.} 
We further assess the effectiveness of the proposed iterative tuning in Fig.~\ref{fig:abl1}(b) and Tab.~\ref{tab:ab_about_model}. It is clearly observed that without iterative tuning, the generated video has the challenge of maintaining temporal coherence (which is marked in \textbf{\textcolor{orange}{orange}} boxes ) in Fig.~\ref{fig:abl1}. This situation worsens when increasing the camera's motion angle. We analyze that the model lacks prior information in specific scenarios (\textit{e.g.}, in the room), which degrades the generation capability of the video inpainting model.



\noindent\textbf{Effectiveness of temporal-packing strategy.} 
In Fig.~\ref{fig:ablation2}, we show the results when lack of a temporal pack strategy during inference. We first generate the video using the camera trajectory $\mathcal{T}^{a}$. Then we ablate the influence of temporal pack strategy using camera trajectory $\mathcal{T}^{b}$. Since there are no extra constraints and conditions, the inpainted area in multi-view video fails to preserve the consistency (which is marked in \textbf{\textcolor{Pink}{pink}} boxes). In contrast, our approach generates the multi-view video with consistent overlap content, which further demonstrates the effectiveness of the proposed strategy. 

\section{Conclusion}
In this paper, we present \ours, a novel 4D video generation framework that reformulates 4D video creation as a video inpainting task, generating more realistic and controllable results with minimal additional training. Specifically, we first generate the composite mask using the dynamic point cloud and double-reprojection strategy. To handle temporal consistency under large camera motion, we design a self-iterative tuning strategy that gradually increases the viewing angles during training. To maintain the multi-view video consistency, the temporal-packing inference is introduced to enhance generation quality. Our method effectively leverages the prior knowledge of the video inpainting model without degrading its original performance, enabling the generation of 4D videos with consistent multi-view coherence. \\
\textbf{Limitations.} The limitation of our method is discussed in the appendix.

\bibliographystyle{plain}
\bibliography{neurips_2025}

\end{document}